# A Dataset and Preliminary Results for Umpire Pose Detection Using SVM Classification of Deep Features

Aravind Ravi*, Harshwin Venugopal*, Sruthy Paul†, Hamid R. Tizhoosh‡
*Department of Systems Design Engineering, University of Waterloo, Canada
†Department of Electrical and Computer Engineering, University of Waterloo, Canada
‡ Kimia Lab, University of Waterloo, Canada [kimia.uwaterloo.ca]
{aravind.ravi, hvenugopal, s25paul, tizhoosh}@uwaterloo.ca

*Abstract*—In recent years, there has been increased interest in video summarization and automatic sports highlights generation. In this work, we introduce a new dataset, called SNOW, for umpire pose detection in the game of cricket. The proposed dataset is evaluated as a preliminary aid for developing systems to automatically generate cricket highlights. In cricket, the umpire has the authority to make important decisions about events on the field. The umpire signals important events using unique hand signals and gestures. We identify four such events for classification namely SIX, NO BALL, OUT and WIDE based on detecting the pose of the umpire from the frames of a cricket video. Pre-trained convolutional neural networks such as Inception V3 and VGG19 networks are selected as primary candidates for feature extraction. The results are obtained using a linear SVM classifier. The highest classification performance was achieved for the SVM trained on features extracted from the VGG19 network. The preliminary results suggest that the proposed system is an effective solution for the application of cricket highlights generation.

## I. INTRODUCTION

Automatic video summarization has gained increased attention in the recent past. Sports highlights generation, movie trailer generation, automatic headlines generation for news are some examples of video summarization. The focus of the present work is sports video summarization in the form of highlights. The highlights of a game provide the summary of important events of that game such as a goal in soccer or a wicket in cricket. It is a challenging task to summarize the highlights from sports videos as these videos are unscripted in nature. An efficient approach can be based on identifying key events from the sports video and use them to automatically generate the highlights.

Among sports, cricket is the most popular game in the world after soccer and has the highest viewership rating. A detailed explanation of the game of cricket can be found in [1]. In the game of cricket, the umpire is the person with the authority to make important decisions about events on the field. The umpire signals these events using hand signals, poses and gestures. This innate characteristic of the cricket video can be leveraged as one approach for solving the problem of cricket highlight generation. Therefore, a system can be developed to detect the unique signals and poses shown by the umpire to automatically generate cricket highlights.

A method for umpire pose detection for generating cricket highlights based on transfer learning is proposed in this work. We explore the use of features extracted from the pre-trained networks such as Inception V3 [19] and VGG19 networks [20] pre-trained on ImageNet dataset. A linear support vector machine (SVM) classifier is trained on the extracted features for detecting the pose of the umpire. A new dataset, SNOW, is introduced in this work and all experiments are performed on this dataset. The system built using this dataset is evaluated on cricket videos for highlights generation.

The paper is organized as follows: In Section II, the background work covering existing techniques will be discussed. Section III introduces the proposed dataset. Section IV outlines the overall system design and methodology for evaluating the proposed dataset as a benchmark for umpire pose detection and highlight generation. The results of the experiments and their analysis are discussed in Section V. Section VI concludes the paper by providing directions for future work.

## II. BACKGROUND

### A. Automatic Highlights Generation

A benchmark database such as TREC Video Retrieval Evaluation (TRECVID) [9] for general video indexing, summarization and retrieval has been used in many studies. In the domain of sports video annotation and summarization, several works such as [10] and [11] have been reported. Automatic summarization of soccer videos has been proposed in [5]. Similar studies for sports such as basketball [6], baseball [7], and tennis [8] have also been reported. More recent work, in the domain of sports video summarization, has been reported in [12] and [13] for the game of cricket.

Prior studies have used event detection in cricket videos as the basis for highlight generation. Hari et al. have proposed a method based on intensity projection profile of umpire gestures for detecting events [2]. A technique based on Bayesian belief networks for indexing broadcast sports video has been proposed in [13]. The use of sequential pattern mining to segment cricket videos into shots and identify the visual content is presented in [4]. Another technique is based on fixing wrist bands on umpires for collecting accelerometer data, and using them to classify the gesture of umpires for labelling the events in a cricket video [3]. Although there has been significant amount of work produced in this domain, benchmark datasets for cricket videos are not available for further development on the existing ideas.

### B. CNN as a Fixed Feature Extractor

Convolutional neural networks (CNNs) have outperformed most of the traditional computer vision algorithms for tasks such as image classification and object detection. A CNN is a combination of a feature extractor and a classifier. The convolutional layers of the CNN are the feature extractors. They learn the representations automatically from the input data. The early layers in the CNN learn more generic features such as shapes, edges, and colour blobs, while the deeper layers learn features more specific to that contained in the original dataset. The last fully connected layers of the CNN use these learned features and classify the data into one of the classes.

The availability of open-source pre-trained models [19] [20] has led to the use of "off-the-shelf" CNN features as complementary information channels to existing handcrafted image features. This technique is popularly known as transfer learning. Transfer learning is defined as the ability of a system to recognize and apply knowledge and skills learned in some domains to new domains [14]. The objective is to learn a good feature representation for the target domain. One approach for transfer learning is to use a CNN, pre-trained on ImageNet database [18], as a fixed feature extractor. In this approach, the last layer is removed, and the rest of the network is used for extracting the features. The features thus obtained can be used to train a linear classifier such as a Linear SVM or softmax classifier for the target dataset.

Prior studies have been successful in applying transfer learning for tasks such as object recognition [16]. Several studies have reported success in using pre-trained CNN features for medical image analysis. Kieffer et al. have evaluated the performance of pre-trained deep features and the impact of transfer learning with a small dataset of histopathology images [21]. Other studies have achieved similar results when attempting to utilize pre-trained network features for medical imaging tasks [17] [22]. From these studies, it can be concluded that pre-trained network features can be used as a primary candidate to evaluate novel datasets in the domain of image recognition.

Our work extends on the idea of identifying events in cricket videos based on detecting the pose of the umpire. In addition to this, we propose the SNOW dataset which comprises of five classes of umpire actions each corresponding to four events such as Six, No Ball, Out, Wide (SNOW) and a no action class is included. We set the benchmark evaluation for image classification based on a linear SVM classifier trained on features extracted from pre-trained networks.

### III. DATASET DESCRIPTION

We have collected images of umpires performing various actions pertaining to events such as "Six", "No Ball", "Out" and "Wide". These images have been obtained from various cricket match videos from YouTube and Google images. The dataset comprises of five classes of data. Four classes belonging to the four actions and one no action class in which the umpire does not

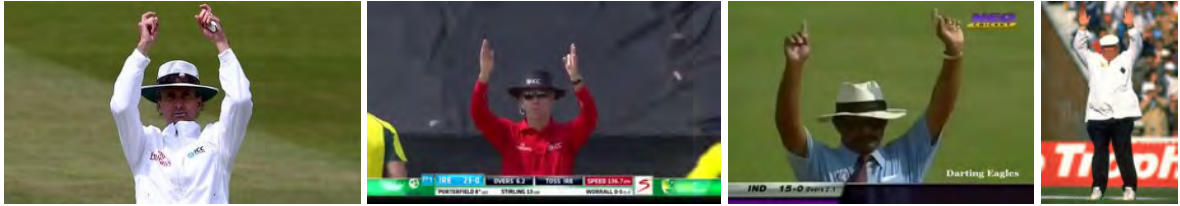

**Six**

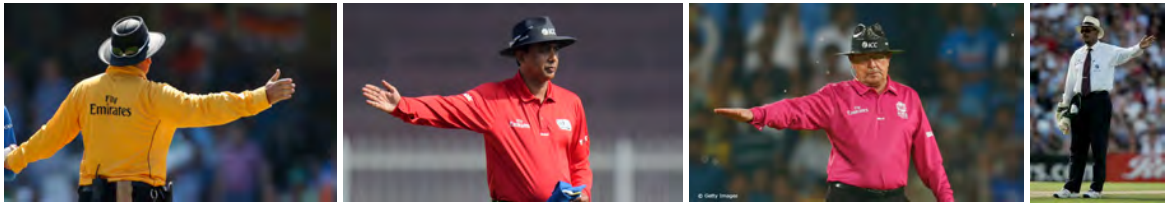

**No Ball**

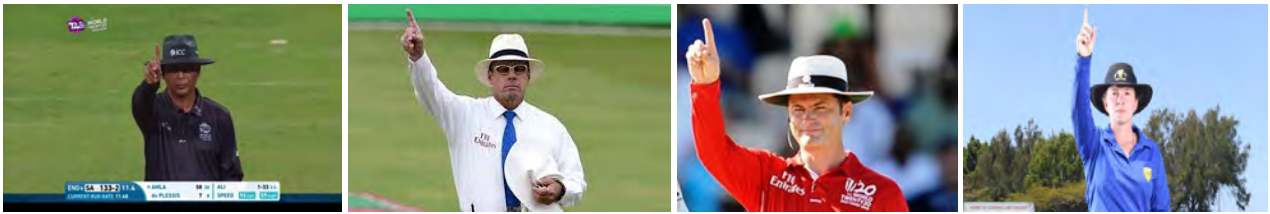

**Out**

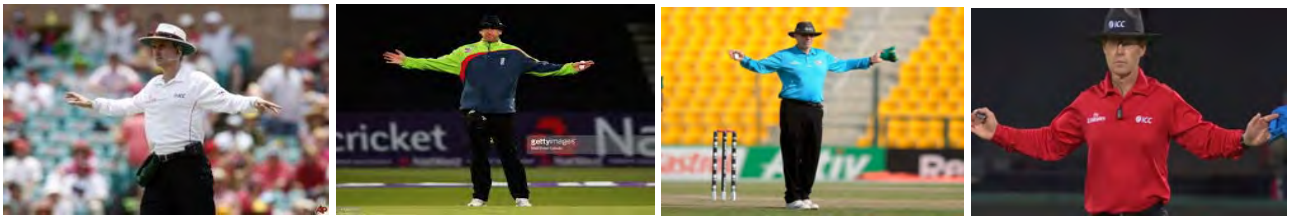

**Wide**

Fig. 1. Examples from the SNOW dataset illustrating umpire images for actions such as "Six", "No Ball", "Out", and "Wide".

perform any action. Each class consists of 78 images summing up to a total of 390 images for all five classes. Fig. 1 illustrates some of the images in the dataset for the four classes of events.

feature extraction, and classification. During the training stage, the input images are pre-processed by performing intensity normalization on the pixel values and resizing to 299 by 299 pixels for the

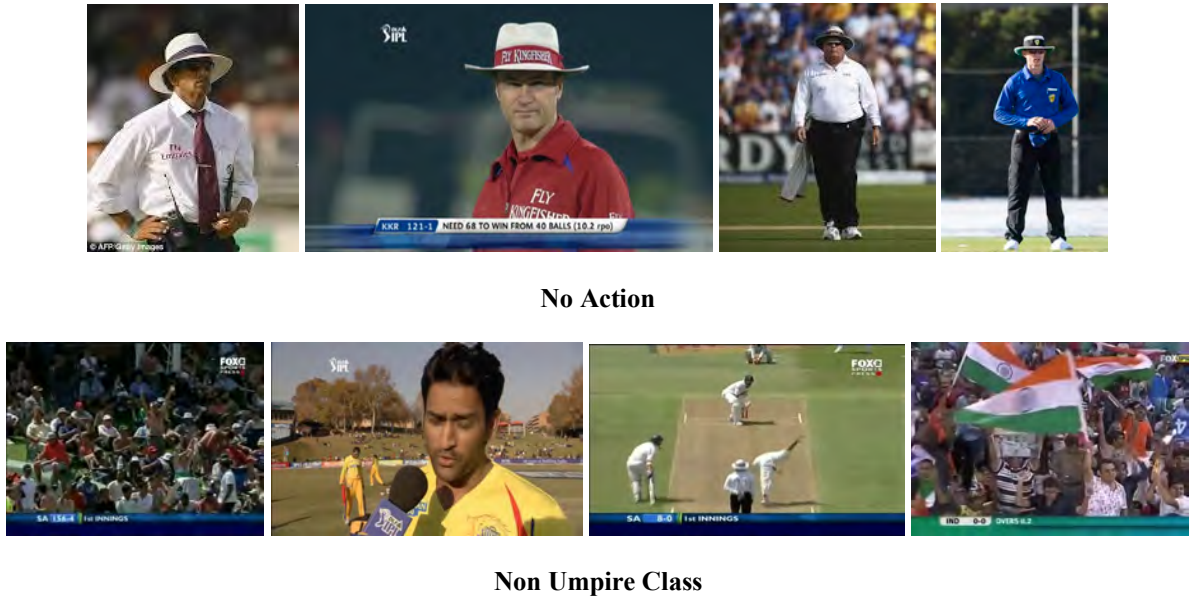

**No Action**

**Non Umpire Class**

Fig. 2. Examples of images for the "No Action" class and "Non Umpire" class.

While this dataset is sufficient for umpire pose detection, there is a need for a non-umpire dataset to distinguish between a frame containing the image of an umpire and that which does not contain an umpire. This non-umpire set contains images of team players, the playing field, crowd, etc. This will facilitate the building of a classification system that can be applied directly on the frames of cricket videos. Fig. 2 shows some examples of images belonging to the no action class and non-umpire class. There is a lot of diversity in the images for every class. There are images shot at different camera angles, orientations and lighting circumstances. Umpire images with different color uniforms, and varying background conditions have been included in this dataset to cover a wide range of game settings. Also, successive frames with slight differences between each other have been extracted from cricket videos and included in the dataset. This dataset has been made available online and can be downloaded [1].

IV. METHODOLOGY

All experiments were performed on Inception V3 and VGG19 pre-trained networks provided in the Keras package for Python [24]. Both networks have been trained on a subset of the ImageNet dataset. The ImageNet dataset is a benchmark dataset for object recognition and was used in ImageNet Large-Scale Visual Recognition Challenge (ILSVRC) [18]. They are trained on more than one million images belonging to 1000 object categories. Therefore, the networks have learned feature representations for a wide variety of categories in the source domain. The Inception V3 consists of 159 layers in total [23]. The VGG19 consists of 26 layers in total with 16 convolutional layers, and 3 fully connected layers.

The proposed system is built in two phases. The first phase involves designing classifiers to distinguish images containing an umpire versus no umpire, and also detect the pose of the umpire, if present. This phase involves the following steps: pre-processing,

Inception V3 network and 224 by 224 pixels for the VGG19 network. The features are extracted from different layers of the pre-trained networks. Finally, these features are used to train a linear SVM classifier to output the class label of the predicted pose of the umpire. The trained classifier models are saved using the python pickle library to be used in the next phase of video summarization.

The second phase involves detecting the events from the cricket videos using the saved classifier models, and generating the summary of the videos. The steps involved in video summarization are: pre-processing, feature extraction, umpire detection, event detection, frame accumulation and video summary generation. The classifier design and the combined pipeline for video summarization are detailed in the following sections.

*A. Phase 1 – Classifier Design*

For the purpose of cricket video summarization two classifiers are designed: Classifier 1 and Classifier 2. The Classifier 1 is designed to distinguish between a frame containing an umpire versus a frame that doesn't contain an umpire. Two sets of data were created. One set containing all 390 umpire images from the SNOW dataset belonging to one class, and a second set containing 390 non-umpire images belonging to the other class. The Classifier 2 is designed for the purpose of umpire pose classification or event detection. This classifier is trained on 390 umpire images from the SNOW dataset belonging to five classes of events such as Six, No Ball, Out, Wide and No Action.

---

[1]Downloading the dataset:
https://drive.google.com/drive/folders/1ljDIz69mJqDzBlUxABP0c34NgdynzWPX

*Experiment 1 – Classifiers based on Inception V3 features*

The steps involved in designing the classifiers are common to both Classifiers 1 and 2. The pre-processing is performed as the first step in processing the images. Features are extracted from the last fully connected layer of the Inception V3 network. The length of the feature vector is 2048. These features, also known as bottleneck features, are extracted for both classifiers. A support vector machine with a linear kernel and regularization parameter C = 10 was empirically identified, based on a grid search between [1,20], to provide the best classification results. Table 1 summarizes the classification performance for both classifiers trained on Inception V3 bottleneck features.

*Experiment 2 – Classifiers based on VGG19 network features*

The approach used in this experiment is similar to the previous one. Pre-processing is followed by feature extraction. Features extracted from the first and second fully connected layers of the VGG19 network are evaluated. The length of the feature vectors for both layers is 4096. Both classifiers are trained on these features. A support vector machine with a linear kernel and regularization parameter C = 10 was used for classification. The classification performance for both classifiers trained using VGG19 features are shown in Table 1.

In all the experiments, the Python package scikit-learn [26] based on the LIBLINEAR [27] implementation was used to train the SVM, and NumPy [28] was used to process and store the data during the experiments. Python's pickle library is used to store and retrieve the trained classification models. The two classifiers are combined into one system for video summarization.

*B. Phase 2 – Video Summarization*

We propose a system to summarize cricket videos by detecting important events that are signaled by the umpire. The saved classification models are combined into one system to realize this. The steps involved are as follows: pre-processing, feature extraction, umpire detection, event detection, frame accumulation and video summary generation.

The input video is processed by extracting the frames sequentially. The frame rate of each video is 25 frames per second (fps). The following steps are performed sequentially for every frame in the video. Each frame is treated as a test image for the classifiers. Intensity normalization is performed as a pre-processing step. Then, the bottleneck features are extracted for these images using the pre-trained networks. These features are first tested on Classifier 1 to detect the presence of an umpire. If the frame is classified as an image belonging to the umpire class, then the features are carried forward, else the frame is discarded and the subsequent frame is processed. In the next step, the features are tested on Classifier 2 to detect the pose of the umpire and detect the event. If the frame was classified as an image belonging to one of the four classes of Six, No Ball, Out or Wide, then the processed frame is accumulated for generating the video summary. If the classified frame belongs to the no action class, then the frame is discarded as this is not relevant for event detection. Frames are accumulated into four individual sequences, each belonging to one of the categories of the detected pose of the umpire. Once all the frames are processed, the accumulated sequences are merged to generate the video summary. Fig. 3 illustrates the overall system design for cricket video summarization.

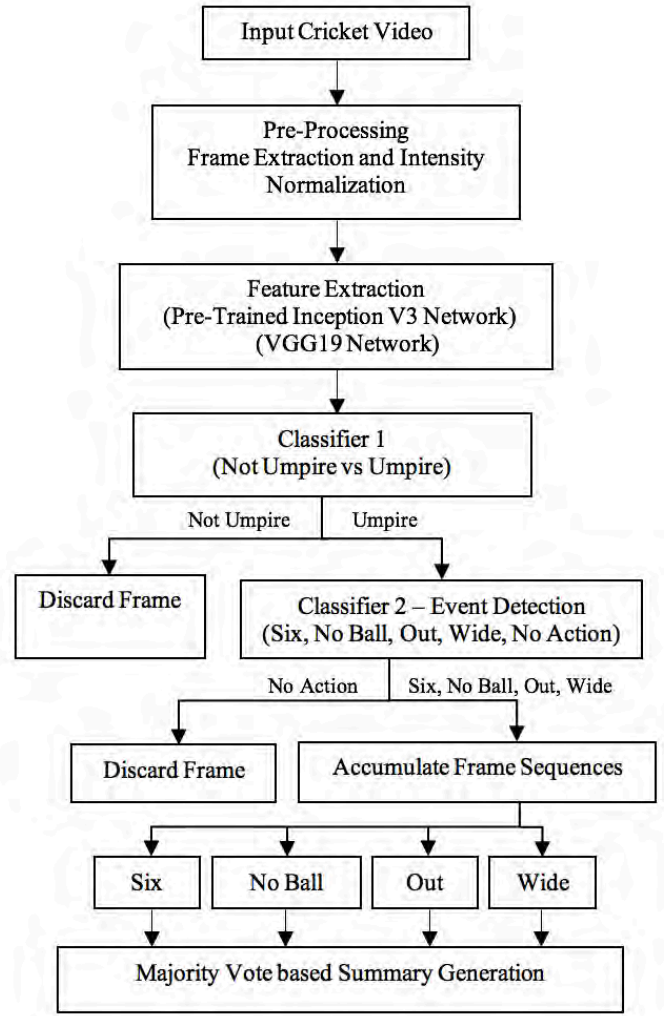

Fig. 3. Overall system design illustrating the steps for video summarization

Processing each frame of a video in sequence is computationally expensive. Also, frame accumulation increases the memory footprint during runtime and is proportional to the length of the video. To overcome this memory overhead, a fixed size buffer is introduced in the frame accumulation and video summary algorithm.

A buffer of size 250 frames is used to accumulate the incoming frames in sequence. For a frame rate of 25 fps this buffer can hold a video clip of 10 second duration. This form of extraction can be viewed as a moving window applied on the original input video. Each frame goes through the entire pipeline of umpire detection followed by umpire pose classification. In this manner, all 250 frames are processed. Based on the final classification result, these frames are accumulated into four individual frame sequences along with the detected class label for each frame. The total number of frames classified into each of the four classes is computed. A majority voter is used to decide the category of the summarized event. For example, if the 250 frames in the buffer contain a scene of an event such as Out, then it is likely to contain frames of an umpire signaling Out. These frames, if correctly classified by Classifier 1 and Classifier 2, are expected to be accumulated more in number into the Out category sequence than other categories. In the ideal case, other category sequences should not contain any frame for a scene containing Out. Based on a majority vote, these 250 frames are merged into a short video to summarize the Out event. The next 250 frames of the input video are processed in a similar manner in subsequent iterations. In a typical cricket video, it can be observed that the events such as Six, No Ball, Out and Wide last a duration of at least 10 seconds. Hence, a moving window of 250 frames was empirically chosen to be ideal to detect and summarize the events.

## V. RESULTS AND DISCUSSIONS

### A. Classifier Design Results

The classifiers are designed based on features extracted from the last fully connected layer of Inception V3 network, the first (fc1) and second (fc2) fully connected layers of the VGG19 network. Classifier 1 and Classifier 2 are trained on 80% of the dataset and the remaining 20% of the dataset is used for testing the classification performance. These classifiers are validated based on a 10-fold cross-validation and Jack-Knife or leave-one-out validation on the training data. The test accuracy is calculated based on the remaining 20% of the unseen data. The classification results are tabulated in Table 1.

Classifier 1 is trained for umpire detection. From the results, it is apparent that Classifier 1 has a good performance accuracy for all three feature extraction methods. The performance of the classifier on features extracted using Inception V3 and fc2 layers of VGG19 are almost similar. The highest accuracy is achieved for the SVM trained on features extracted from fc1 layer of the VGG19 network.

Fig. 4 illustrates some examples of correct classifications and misclassifications by Classifier 1 trained on the fc1 layer features. Although the exact reason for misclassification is not known, analyzing certain features in the images can provide some insight. In Fig. 4B, the image on the left contains a player wearing a hat that is similar to the one found in umpire images. The image on the right, shows a player whose action resembles that of an umpire signaling a WIDE. Therefore, it can be inferred that these features could have led to the misclassifications. Nevertheless, it can be observed that the proposed dataset and feature extraction methods have been effectively used in detecting umpire images.

TABLE I. TRAINING AND TEST ACCURACIES OF CLASSIFIER 1 AND CLASSIFIER 2

| CLASSIFIER | FEATURES | ACCURACIES | | |
|---|---|---|---|---|
| | | 10-FOLD | JACK-KNIFE | TEST |
| 1 | INCEPTION V3 | 96.97% | 97.76% | 94.23% |
| | VGG19 – FC1 LAYER | 97.75% | 97.59% | 96.15% |
| | VGG19 – FC2 LAYER | 96.47% | 96.79% | 94.87% |
| 2 | INCEPTION V3 | 77.71% | 77.56% | 85.90% |
| | VGG19 – FC1 LAYER | 82.43% | 81.09% | 83.33% |
| | VGG19 – FC2 LAYER | 78.14% | 81.09% | 78.21% |

Classifier 2 is trained on the five class SNOW dataset for umpire pose detection. The highest performance is achieved for the SVM trained on features extracted from the fc1 layer of the VGG19 network. This classifier was tested on novel images that are not part of the training data. Fig. 5 shows some examples of correct detections and misclassifications in umpire pose detection.

Fig. 5A, illustrates images that are correctly classified. In Fig. 5B, some misclassified examples are shown. This could be due to the following reasons: on the left is an image of a player whose action resembles that of an umpire signaling a Six, and on the right it resembles that of an umpire signaling a No Ball. It can be concluded that Classifier 2 is successful in classifying any image into one of the five classes of poses. But it lacks the ability to detect the presence of an umpire in the image as it has not been designed to do so. Hence Classifier 1 is used in the overall system pipeline to add the ability of filtering the umpire images from non-umpire images.

Similar performance is observed for classifiers trained on bottleneck features extracted from Inception V3 network. Classifiers 1 and 2 are combined such that both are trained using the same feature extraction method. For example, Classifier 1 trained on features extracted using fc1 layer of VGG19 is combined with the Classifier 2 trained using the same feature extraction method. The combined system is evaluated for its effectiveness in summarizing the cricket video.

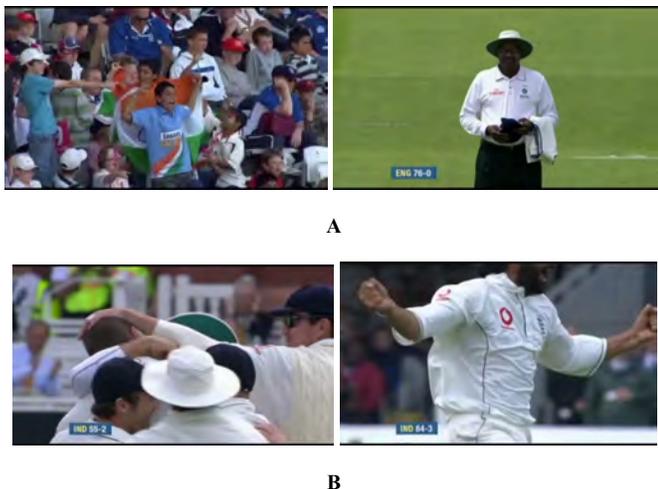

A

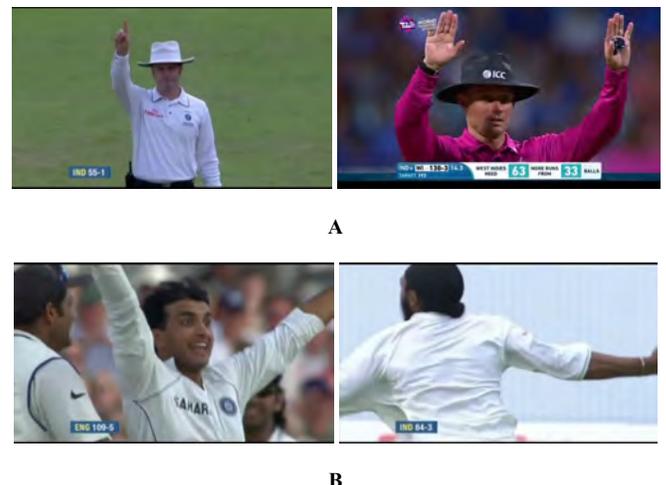

A

B

Fig. 4A. Classifier 1 correctly classifying novel images as non-umpire (left) and umpire (right). Fig. 4B. Classifier 1 misclassifying both novel images as umpire images.

B

Fig. 5A. Classifier 2 correctly classifying novel images as OUT (left) and Six (right). Fig. 5B. Classifier 2 misclassifying both non umpire images as Six (left) and No Ball (right).

*B. Cricket Video Summarization Results*

The proposed cricket video summarization system is assessed for its ability to effectively generate the highlights on novel test videos. The videos used for this purpose are based on a collection of cricket videos available on YouTube. Two short video segments, V1 and V2, containing events signaled by the umpire are synthesized from the collection of videos. V1 consists of 8 events and V2 consists of 5 events with a total of 13 events is considered.

The summarization performance is measured based on the metrics of True Positive Rate (TPR) and Positive Prediction Value (PPV). TPR measures the ratio between correctly summarized events and actual number of events. PPV measures the ratio between correctly summarized events and all summarized events. They are defined as:

$$TPR = \frac{TP}{TP+FN}, \quad (1)$$

$$PPV = \frac{TP}{TP+FP}, \quad (2)$$

where a true positive (TP) is reported if the system summarizes an event correctly as the actual event. A false positive (FP) is reported if the system summarizes an event as belonging to one of the four categories when the actual event is not present in the video. If the system fails to detect an event, it is recorded as a false negative (FN). In the ideal case, a good performing system is required to have a high TPR and PPV. Table 2 presents the results of cricket video summarization tested on the two synthesized cricket videos. The system achieves very high TPR of 0.9166 and PPV of 0.9166 for both sets of feature extraction methods.

TABLE II. CRICKET VIDEO SUMMARIZATION TESTED ON TWO SYNTHESIZED CRICKET VIDEOS

| VIDEO | EVENT CATEGORY | ACTUAL NUMBER OF EVENTS | FEATURES | | | | | |
|---|---|---|---|---|---|---|---|---|
| | | | INCEPTION V3 | | | VGG19 - FC1 | | |
| | | | TP | FP | FN | TP | FP | FN |
| V1 | SIX | 1 | 1 | 0 | 0 | 1 | 0 | 0 |
| | NO BALL | 3 | 2 | 0 | 1 | 2 | 0 | 1 |
| | OUT | 2 | 1 | 1 | 0 | 1 | 1 | 0 |
| | WIDE | 2 | 2 | 0 | 0 | 2 | 0 | 0 |
| V2 | OUT | 5 | 5 | 0 | 0 | 5 | 0 | 0 |
| | TOTAL | 13 | 11 | 1 | 1 | 11 | 1 | 1 |
| TRUE POSITIVE RATE | | | 0.9166 | | | 0.9166 | | |
| POSITIVE PREDICTION VALUE | | | 0.9166 | | | 0.9166 | | |

Fig. 6. illustrates some of the frames that caused the FP and FN. In all cases, Classifier 1 has detected the umpire correctly. The misclassifications have been caused by the Classifier 2, which is responsible for detecting the pose of the umpire. The Classifier 2

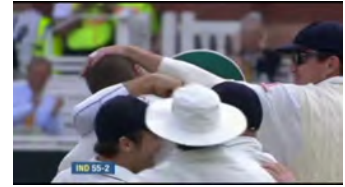

A

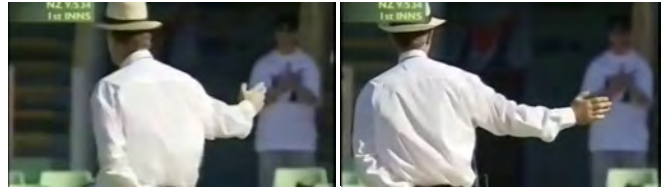

B

Fig. 6A. False Positive – Umpire image containing no action, but falsely classified as an Out. Fig. 6B. False Negative – Umpire signaling a No Ball, but the frame is falsely classified as No Action (left) and the subsequent frame correctly classified as a No Ball (right)

is expected to classify the frame shown in Fig. 6A into the no action class, but the frame has been classified as Out. Similarly, the frame on the left of Fig. 6B has been misclassified as no action class, while the umpire is signaling a No Ball. In contrast, the subsequent frame from the same video has been classified correctly as a No Ball. Therefore, two issues can be identified: the FPs can be attributed to the performance of Classifier 2, and the FNs can be attributed to the majority voting scheme of the video summarization technique. In the case of the FN, it is observed that more number of umpire frames are classified as belonging to the no action class when compared to the number of frames correctly classified as a No Ball. Hence, the system has discarded these frames from the buffer and has not included them in the final summary of the video. To achieve better performance, alternate algorithms can be explored.

It is important to observe the time complexity of the system while evaluating its performance. The time taken for each subcomponent in the proposed system is analyzed. The results are calculated based on experiments performed on a machine with Intel Core-i5 CPU and 8 GB RAM. Table 3 lists the observations of the time taken for feature extraction and training the SVM classifiers.

TABLE III. TIME TAKEN FOR FEATURE EXTRACTION AND TRAINING THE SVM FOR CLASSIFIERS 1 AND 2

| CLASSIFIER | FEATURES | TIME (S) | |
|---|---|---|---|
| | | FEATURE EXTRACTION | TRAINING |
| 1 | INCEPTION V3 | 386.28 | 9.47 |
| | VGG19 – FC1 LAYER | 551.93 | 6.42 |
| 2 | INCEPTION V3 | 132.01 | 30.64 |
| | VGG19 – FC1 LAYER | 237.90 | 19.40 |

The training time for the SVM trained on features of fc1 layer of VGG19 network is less than time taken to train on the bottleneck features of Inception V3 network. The length of the features extracted on the fully connected layers of VGG19 and Inception V3 are 4096 and 2048 respectively. A reasonable explanation for reduced training time could be that the SVM classifier converges faster in a higher dimensional (4096) feature space when compared to the 2048-dimensional feature space.

Among the two synthesized videos, V1 is the longest video of 2.5-minutes duration. The combined pipeline using VGG19 features takes approximately 57.7 minutes to summarize all the events, and 47.5 minutes for features extracted using Inception V3. The increased time complexity can be attributed to the fact that each frame of the video is processed sequentially. This can be improved by segmenting the original video based on a fixed buffer size and process the segments in parallel. Also for longer duration videos, big data frameworks such as Hadoop or Spark can be employed [29] [30].

## VI. CONCLUSIONS

We proposed a system capable of summarizing cricket videos in the form of highlights based on detecting important events from the pose of the umpire. A new dataset, SNOW, containing umpire images for events such as Six, No Ball, Out and Wide was introduced in this work. This dataset has been made publicly available.

The classification results indicated that features extracted from pre-trained networks such as VGG19 and Inception V3 provide a good performance baseline for this dataset. The combined system has been tested on cricket videos and is successful in detecting the majority of the events present in the video. The preliminary results obtained for the SNOW dataset suggest that this dataset is effective for the application of cricket highlights generation. Improvements can be made in the areas of classifier design and summarization technique to minimize false positives and false negatives. Features extracted from earlier parts of a pre-trained network and other feature extraction techniques can be explored. Timing overhead can be improved by extending the design to include concepts like parallelization and distributed computing.

One may also compare the deep features against recently introduced handcrafted features that seem to be quite accurate when combined with classifiers like SVM [31].